\def\year{2020}\relax
\documentclass[letterpaper]{article} 
\usepackage{aaai20}  
\usepackage{times}  
\usepackage{helvet} 
\usepackage{courier}  
\usepackage[hyphens]{url}  
\usepackage{graphicx} 
\usepackage{amsfonts}
\usepackage{graphicx}  
\usepackage[
  separate-uncertainty = true,
  multi-part-units = repeat
]{siunitx}
\usepackage{amsmath, amsthm, amssymb}
\usepackage{booktabs}
\usepackage{algorithm,algpseudocode}
\usepackage{wrapfig}

\usepackage{hyperref}
\hypersetup{
colorlinks=true,
    citecolor = {blue}
}

\usepackage{xcolor}

\title{Meta-CoTGAN: A Meta Cooperative Training Paradigm for Improving  Adversarial Text Generation}

\author{Haiyan Yin, Dingcheng Li, Xu Li, and Ping Li\\
Cognitive Computing Lab\\ Baidu Research\\
 No.10 Xibeiwang East Road, Beijing, 10085, China\\
10900 NE 8th St. Bellevue, WA 98004, USA\\
haiyanyin18@gmail.com, \{lidingcheng, lixu13\}@baidu.com, \ pingli98@gmail.com\\
}
	
\begin{document}

\maketitle

\begin{abstract}
Training generative models that can generate high-quality text with sufficient diversity is an important open problem for Natural Language Generation (NLG) community. Recently, generative adversarial models have been applied extensively on text generation tasks, where the adversarially trained generators alleviate the \textit{exposure bias} experienced by conventional maximum likelihood approaches and result in promising generation quality.
However, due to the notorious defect of \textit{mode collapse} for adversarial training, the adversarially trained generators face a quality-diversity trade-off, i.e., the generator models tend to sacrifice generation diversity severely for increasing generation quality.
In this paper, we propose a novel approach which aims to improve the performance of adversarial text generation via efficiently decelerating \textit{mode collapse} of the adversarial training.
To this end, we introduce a cooperative training paradigm, where a language model is cooperatively trained with the generator and we utilize the language model to efficiently shape the data distribution of the generator against \textit{mode collapse}. Moreover, instead of engaging the cooperative update for the generator in a principled way, we formulate a meta learning mechanism, where the cooperative update to the generator serves as a high level meta task, with an intuition of ensuring the parameters of the generator \textit{after} the adversarial update would stay resistant against \textit{mode collapse}. In the experiment, we demonstrate our proposed approach can efficiently slow down the pace of \textit{mode collapse} for the adversarial text generators. Overall, our proposed method is able to outperform the baseline approaches with significant margins in terms of both generation quality and diversity in the testified domains.
\end{abstract}

\section{Introduction}
Generative models are trained to learn the true data distribution from the training set and are capable of generating new data points when the training is completed. In recent years, they have been successfully applied to a wide range of applications, including image generation~\cite{arjovsky2017wasserstein}, stylization~\cite{ulyanov2016instance}, semi-supervised classification~\cite{radford2015unsupervised}, and natural language generation~\cite{bahdanau2014neural,li2019multi,sun2018logician}, etc. In this paper, we tackle the emerging task of text generation, which is typically modeled as a sequential discrete data generation process \cite{lamb2016professor}. Such tasks play a pivot role in many real world applications, such as machine translation~\cite{sutskever2014sequence}, text summarization~\cite{li2017deep,liu2012query}, and dialogue systems~\cite{wen2015semantically,li2016deep}.

The training of sequential text generation models has been greatly relying on applying \textit{teacher forcing} over autoregressive models, i.e., optimizing with maximum likelihood estimation (MLE)~\cite{che2017maximum}. However, training the generative models with \textit{teacher forcing} would suffer from \textit{exposure bias}~\cite{rajeswar2017adversarial}, i.e., the models are fed to their predicted data rather than the ground-truth data at inference time and thus result in generating poor samples due to the accumulated error.
To address the \textit{exposure bias} issue, a major on-going research for text generation centers on utilizing adversarial training techniques to derive better text generation models. Generally, such attempts could be classified into the following two strands~\cite{chen2018adversarial}: the first line of approaches combine generative adversarial network (GAN)~\cite{goodfellow2014generative} with reinforcement learning (RL), denoted as RL-based; the second line of approaches solely play the two-player adversarial game without using RL, denoted as RL-free.

Both RL-based and RL-free text generation approaches suffer from \textit{mode collapse}, a notoriously known challenge for training GAN-based models~\cite{arjovsky2017wasserstein}.
That is, as the adversarial training progresses, the generated distribution tends to contrast towards generating subset of modes for the data. As a result, the generator outputs repeated sentences and thus no longer expressively represents the data generating distribution. Such effect has been quantitatively evaluated in a recent study, which shows that the entropy of the generator's output distribution would experience a clear drop when moving from MLE training to adversarial training phase~\cite{caccia2018language}. To derive better text generation models with GAN-based techniques, one critical thing is to achieve a better quality-diversity trade-off by efficiently slowing down the \textit{mode collapse} of the adversarial generator, i.e., to let the generator get abundant gradient information from adversarial update for making its output more real (i.e., improve quality) while bearing with small \textit{mode collapse} effect (i.e., decrease diversity). However, limited number of existing RL-based or RL-free approaches explicitly consider dealing with \textit{mode collapse} of GAN training. In this work, we propose a cooperative training mechanism which explicitly tackles the challenge of \textit{mode collapse} for adversarial training, resulting in an improved text generation model.

Overall, the contributions of this paper are three-fold. Firstly, we propose a novel cooperative training approach where we utilize a language model to efficiently shape the output distribution of the adversarial text generator. Our proposed approach could efficiently slow down the \textit{mode collapse} of the adversarial text generator and thus lead the text generation towards a better quality-diversity trade-off. Secondly, to optimize the cooperative training loss for the generator, we propose a novel meta-learning mechanism. In our setting, the cooperative training task serves as a meta task and the adversarial training serves as a base task. Thus, our proposed approach ensures that the generator parameters \textit{after} the adversarial update would be resistant for \textit{mode collapse}. Thirdly, we conduct extensive experiments on synthetic and real-world datasets to demonstrate that our proposed approach is able to produce better text generation models in terms of both the quality and the diversity.

\section{Related Work}

Besides the conventional approaches of training language models with \textit{teacher forcing}, today's approaches for text generation could be generally classified as RL-based or RL-free approaches. Most RL-based approaches formulate text generation as a Markov Decision Process (MDP). Often, the generator is updated by policy gradient algorithm~\cite{sutton2000policy} or its variants using reward signals derived from GAN's discriminator. Prominent examples for this type of approaches include SeqGAN~\cite{yu2017seqgan}, RankGAN~\cite{lin2017adversarial}, LeakGAN~\cite{guo2018long} and MaskGAN~\cite{fedus2018maskgan}. The noisy reward signals derived from the discriminator model makes such RL-based models suffer from high-variance gradients to update the generator's parameters.
Besides high-variance of gradient, the RL-based approaches also face the difficulties brought by partial sequence evaluation, slow learning, and sensitive hyperparameters~\cite{caccia2018language}. Considering such challenges for the RL-based approaches, in this work, our proposed method resides in, but not restricted to, the category of RL-free approach for text generation.
Prominent examples of RL-free approaches include TextGAN~\cite{zhang2017adversarial}, FM-GAN~\cite{chen2018adversarial}, GSGAN~\cite{kusner2016gans}, and RelGAN~\cite{nie2018relgan}. Such approaches feed the generator with low variance gradient and often lead to more stable training.

Most of the adversarial text generation models are firstly pretrained by MLE, and then are continuously optimized by adversarial training under either RL-based or RL-free mechanism. When switched from MLE training to adversarial training phase, the generator models for both RL-based and RL-free approaches would suffer from \textit{mode collapse} issue.
In this work, our core intuition is to utilize a cooperatively trained language model to decelerate the \textit{mode collapse} of adversarial training. Such intuition of utilizing language model to facilitate adversarial text generation aligns with the works proposed in~\cite{xu2018dp,lu2018neural}. In ~\cite{xu2018dp}, the discriminator for adversarial training is modeled as a language model, which maximizes the probability for real data and minimizes that for generated data. Furthermore, the output derived from the language model is adopted as reward signal to promote generation diversity under an RL-based set-up. Our work is mostly related to the cooperative training method proposed in~\cite{lu2018cot}, where a language model is trained online to offer a target distribution for minimizing the Jensen-Shannon divergence between the real data distribution and the generated distribution. In our work, we adopt a similar strategy  to train the language model, but the cooperative training for the generator model is different from~\cite{lu2018cot}. Furthermore, we propose a distinct meta learning set-up to optimize the cooperative training loss for the generator. To the best of our knowledge, our work is the first attempt that adopts meta learning on text generation GANs.

\section{Preliminaries}
The task of text generation is typically modelled as sequential discrete data generation process. Let $\{x_j\}_{j=1}^N$ be the $N$ data points drawn from an underlying data generating distribution $p_{data}$. Each data point is represented as a sequence of discrete tokens: $\textbf{x}= (y_1, ..., y_T)$, where $y_i$ denotes the $i$-th token and $T$ denotes the length of the sequence.
Let $G_\theta$ denote the generator model parameterized by $\theta$. Conventional text generation approaches typically train a language model with maximum likelihood estimation (MLE) as follows:
\begin{equation}\notag
\min\limits_{\theta} \, \mathop{\mathbb{E}}_{\textbf{x}\sim p_{data}} \: [-\mbox{log} \,G_\theta (\textbf{x})],
\end{equation}
where the probability of each sequence $\textbf{x}$ is represented in an autoregressive manner:
\begin{equation}\notag
G_\theta(\textbf{x}) = \prod_{i=1}^T \,G_\theta\,(y_i\,|\,y_{<i};\theta),
\end{equation}
with $y_{<i}$ denoting the sequence of previous tokens $y_1, ..., y_{i-1}$.

The approaches utilizing GANs for text generation attempt to play a two-player game between the generator $G_\theta$ and a discriminator $D$. Let the discriminator $D$ be parameterized by $\phi$.  Under the adversarial set-up, the generator $G_\theta$ is trained to generate realistic sentences given samples from $p_{data}$, and the discriminator $D_\phi$ attempts to distinguish between $G_\theta$'s generating distribution $p_\theta$ and the real data distribution $p_{data}$. Thus, the above mentioned process could be formulated as an adversarial training mechanism as follows:
\begin{align}\label{eq:adv}
\min\limits_{\theta} \max\limits_{\phi} \mathop{\mathbb{E}}_{{\textbf{x}}\sim p_{data}}[\mbox{log}(D_{\phi}({\textbf{x}})]  + \mathop{\mathbb{E}}_{{\textbf{x}}\sim p_\theta} [\mbox{log}(1 - D_\phi(G_\theta({\textbf{x}}))],
\end{align}
where the generator $G_\theta$ and discriminator $D_\phi$ attempt to minimize and maximize the function, respectively. We denote the adversarial loss in (\ref{eq:adv}) in terms of the generator model and the discriminator model as $\mathcal{L}_{adv}(\theta)$ and $\mathcal{L}_{adv}(\phi)$, respectively.

With the autoregressive generation process, the $i$-th token $y_i$ is generated by sampling from the generator's output distribution, conditioned on its previous tokens $y_{<i}$. Performing such sampling introduces considerable difficulty for the generator to utilize the discriminator's prediction outcome. That is,  the backpropagation route for adversarial loss, i.e.,
\begin{equation}\notag
\frac{\partial \, \mathcal{L}_{adv}}{\partial \, \theta} = \sum_{i=0}^{T-1} \frac{\partial \, \mathcal{L}_{adv}}{\partial \, y_{t+1}} \frac{\partial \, y_{t+1}}{\partial \, \theta},
\end{equation}
becomes non differentiable w.r.t. the generator's parameters $\theta$, since $\frac{\partial \, y_{t+1}}{\partial \, \theta}$ would be zero due to the sampling. To overcome the above issue, the RL-based approaches mostly rely on the REINFORCE algorithm~\cite{williams1992simple} or its variants to derive the gradient to optimize the generator, where the discriminator's predictions could be utilized to derive reward signals. The RL-free approaches often relax the non-differentiable sampling function by some continuous approximations, such as \textit{soft-argmax}~\cite{zhang2017adversarial} or \textit{gumbel-softmax}~\cite{jang2016categorical}. In this paper, our proposed approach adopts the \textit{gumbel-softmax} relaxation which models the effect of sampling as introducing noise to the input so that the outputs become continuous and differentiable. Specifically, the noise is modeled by a Gumbel distribution, which is formed as follows:
\begin{align*}
g_t^{(i)} = -\mbox{log}\,(-\mbox{log}\,(U_t^{(i)})), \:\: \mbox{with} \: U_t^{(i)} \sim \mbox{Uniform} (0,1),
\end{align*}
where $g_t^{(i)}$ denotes the Gumbel noise to be applied to the $i$-th logits. With the Gumbel noise, the token for next step $y_{t+1}$ is derived in a deterministic manner:
\begin{align}\notag
y_{t+1} = \mbox{one\_hot}\,\big(\mathop{\mbox{arg max}}_{1 \leq i \leq V}(o_t^{(i)} + g_t^{(i)})\big),
\end{align}
where $\textbf{o}_t \in \mathbb{R}^V$ denotes the logits output by the generator for sampling token $y_{t+1}$, and $V$ denotes vocabulary size.
To make the discriminator's loss differentiable, the \textit{argmax} operator is replaced by a \textit{softmax} function $\sigma(\cdot)$, i.e., $\hat{\textbf{y}}_{t+1} = \sigma(\beta(\textbf{o}_t + \textbf{g}_t))$, where $\beta$ is a real-valued temperature hyperparameter, with $\beta > 0$.

\section{Methodology}
Language generators trained with adversarial training mechanism (both RL-based and RL-free approaches) suffer from \textit{mode collapse} when switched from \textit{teacher forcing} to the adversarial training phase.
In this section, we introduce a novel meta cooperative training algorithm to overcome such challenges. Overall, our objective is to achieve a better quality-diversity trade-off for the language generators via decelerating \textit{mode collapse} of their adversarial training. That is, the algorithm allows the generator to get abundant gradient information from the adversarial training for increasing generation quality, while sacrificing little in terms of generation diversity.
Overall, we engage a language model to decelerate the \textit{mode collapse} of the generator's output distribution. The language model is cooperatively trained with the generator $G_\theta$ during adversarial training. We utilize the output of language model over samples from real data distribution $p_{data}$ to shape the generator's output distribution.
Furthermore, the supervision is formulated with a meta optimization setup.

\begin{figure}[t!]
\begin{center}
\includegraphics[width=0.8\columnwidth]{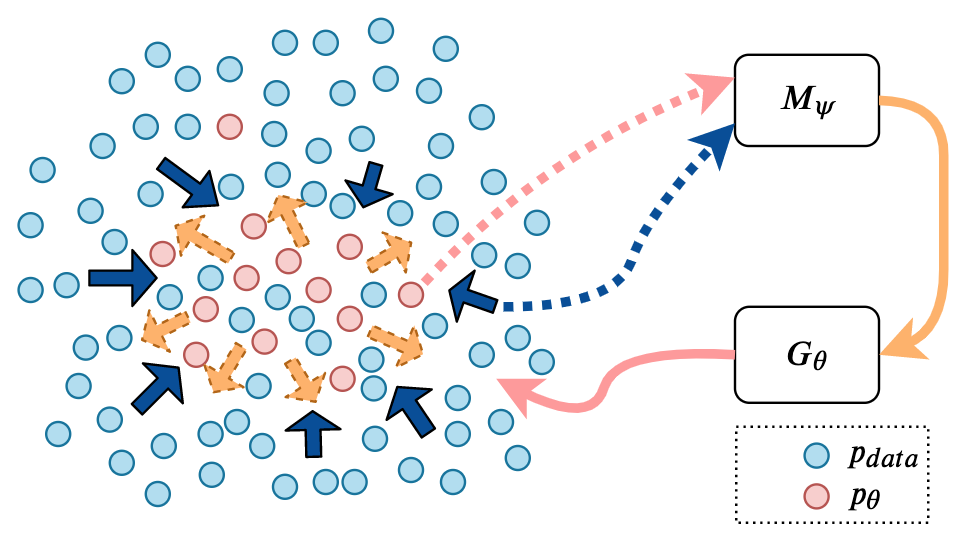}
\end{center}
\vspace{-0.15in}
\caption{Depiction for our proposed cooperative training mechanism. The generator trained with adversarial training tend to suffer from \textit{mode collapse} (dark blue arrows). We engage a language model to supervise the data distribution of $G_\theta$ to decelerate \textit{mode collapse} (yellow arrows). The language model is trained from a mixtured distribution of samples from $p_\theta$ and $p_{data}$. The supervision from language model to the language generator works on samples from $p_{data}$. The generator is updated by the adversarial loss and the cooperative training loss.
}
\label{fig:overview}
\end{figure}

\subsection{Cooperative Training Formulation} 
We introduce a cooperative training paradigm that engages an interleaved training procedure for an adversarial generator $G_\theta$, an adversarial discriminator $D_\phi$, and a language model $M_\psi$, where $\psi$ denotes the parameters for the language model. Figure~\ref{fig:overview} depicts a high-level overview for the proposed cooperative training procedure. When the generator $G_\theta$ is trained by the adversarial loss, its generation diversity would progressively decrease for increasing the generation quality due to \textit{mode collapse} issue. To overcome such challenge, we cooperatively train a language model $M_\psi$. The language model would pose a supervision over $G_\theta$'s output distribution towards preserving desirable generation probability for the real data.

During the cooperative training process, the language model is optimized consistently by MLE loss. To offer a smoothly changing target distribution for the generator, it is trained with data from a mixture distribution with balanced samples from real data and generated data, i.e., $\frac{1}{2} (p_{data} + p_\theta)$.
Formally, the cooperative training loss $\mathcal{L}_{cot}(\psi)$ for updating the language model with MLE is defined in (\ref{eq:cot}). It could be interpreted as minimizing the direct KL divergence between $M_\psi$ and an optimal mixture density model $M^*$ which has a distribution of $\frac{1}{2} (p_{data} + p_\theta)$.
\begin{align}\notag
& \nabla_\psi \: \mathcal{L}_{cot} (\psi) \\\notag
  =&  -  \frac{1}{2}\:\nabla_\psi \Big(\mathop{\mathbb{E}}_{\textbf{x}\sim p_\theta} \mbox{log}\big(M_\psi(\textbf{x}) \big)
+ \mathop{\mathbb{E}}_{\textbf{x}\sim p_{data}}\mbox{log}\big(M_\psi(\textbf{x})\big)\Big)\\\label{eq:cot}
  =&  \nabla_\psi \mathop{\mathbb{E}}_{\textbf{x}\sim M^*} \Big( \mbox{log} \frac{M^*(\textbf{x})}{M_\psi(\textbf{x})}\Big) 
  =  \nabla_\psi \:\: \mbox{KL}\:(M^* || M_\psi),
\end{align}

Consistently updating the language model $M_\psi$ with samples from real data and using the \textit{teacher forcing} loss makes it experience mild \textit{mode collapse} effect. Thus, its output predictions could offer an effective supervision over the generator $G_\theta$'s output distribution for decelerating \textit{mode collapse}. Moreover, updating $M_\psi$ with the mixture distribution, compared to only using the real data distribution, would offer a target distribution that is smoothly changing towards the generator's update, which turns out to be more beneficial.
Formally, the cooperative training loss for the generator model is proposed as follows,
\begin{align} \notag
 \mathcal{L}_{cot} (\theta) = &  \: \mbox{KL} \big( M_\psi (\textbf{x}) \:|| \: G_\theta (\textbf{x}) \big) \\\label{eq:cot_gen_loss}
 = & \sum_{i=1}^T M_\psi(y_i) \mbox{log} \frac{M_\psi(y_i) }{G_\theta(y_i) },
\end{align}
where $y_i$ is the $i$-th token from the sequence $\textbf{x}$. Thus, the KL-loss distills the output distribution given by the language model to the generator~\cite{hinton2015distilling,rusu2015policy,yin2017knowledge}. When considering the \textit{mode collapse}, we would only be interested in preserving the distribution for the real data from $p_{data}$, rather than those from $p_{\theta}$. Therefore, when optimizing (\ref{eq:cot_gen_loss}), we only adopt samples from the real data distribution $p_{data}$ to compute the KL-loss.
With the above cooperative training loss, the gradient for updating the generator's parameters is derived as follows,
\begin{align} \notag
  \nabla_\theta  \: \mathcal{L}_{cot} \:(\theta)
 & =  \nabla_\theta  \: \Big( \sum_{i=1}^T M_\psi(y_i) \mbox{log} \frac{M_\psi(y_i) }{G_\theta(y_i) } \Big) \\\notag
 & = - \sum_{i=1}^T  M_\psi(y_i)  \nabla_\theta \: \mbox{log}(G_{\theta}(y_i)).
\end{align}
As such, the effect of applying cooperative training on the generator is equivalent to increasing the density of the real data in a weighted manner.

\subsection{Meta Cooperative Optimization}
In this section, we introduce a meta learning paradigm to interleave the optimization of the adversarial training loss $\mathcal{L}_{adv}(\theta)$ and the cooperative training loss $\mathcal{L}_{cot}(\theta)$ for the generator model parameters. Unlike the conventional meta-learning approaches that work on achieving faster learning~\cite{finn2017model}, task generalization~\cite{li2018learning} or deriving adaptive models~\cite{al2017continuous}, our intuition is to preserve the generative distribution for the adversarial text generator model to decelerate its \textit{mode collapse}.


To this end, optimizing the adversarial loss $\mathcal{L}_{adv}(\theta)$ is modelled as a base task, and optimizing the cooperative training loss $\mathcal{L}_{cot}(\theta)$ is modeled as the meta task. With such setting, the meta optimization scheme ensures that after optimizing the generator parameters $\theta$ with the adversarial training loss $\mathcal{L}_{adv}(\theta)$ for increasing generation quality, the resultant parameters would demonstrate considerable resistance towards \textit{mode collapse}, i.e., increasing generation quality while preserving considerable generation diversity.

Formally, we first perform one gradient update on the generator parameters $\theta$ by optimizing the base task loss:
\begin{equation}\notag
\theta'   = \theta - \alpha \, \nabla_\theta \, \mathcal{L}_{adv} (\theta).
\end{equation}
Then, we obtain new samples from the real data distribution: $\textbf{x} \sim p_{data}$ and inference the meta-loss $\mathcal{L}_{cot}(\theta')$ for the real samples on the updated parameters $\theta'$.
The meta gradient is weighted by $\lambda > 0 $ and added to the base task gradient to update the parameters $\theta$. Finally, the adversarial update under our proposed meta cooperative training paradigm could be formulated as below:
\begin{equation*} \label{eq:meta}
\begin{split}
\mathcal{L}_D & = \max\limits_{\phi} \mathop{\mathbb{E}}_{\substack{\textbf{x}_r\sim p_{data}, \\ \textbf{x}_y\sim p_{\theta}}} \mathcal{L}_{adv}(\phi) \\
\mathcal{L}_G & = \min\limits_{\theta}
\mathop{\mathbb{E}}_{\substack{\textbf{x}_r\sim p_{data}, \\ \textbf{x}_y\sim p_{\theta}}} \Big(\mathcal{L}_{adv} (\theta )  + \lambda\mathcal{L}_{cot} (\theta')\Big)\\
\mathcal{L}_M & = \min\limits_{\psi} \mathop{\mathbb{E}}_{\substack{\textbf{x}_r\sim p_{data}, \\ \textbf{x}_y\sim p_{\theta}}} \mathcal{L}_{cot}(\psi)
\end{split}
\end{equation*}
The full algorithm for meta cooperative training is presented in Algorithm~\ref{algo}.

\begin{algorithm}[h!]
\caption{Meta Cooperative Training}
\label{algo}
\begin{algorithmic}[1]
\Require{$G_\theta, D_\phi, M_\psi$, learning rate $\alpha/\beta/\gamma$, training data distribution $p_{data}$}
\Ensure{Generator $G_\theta$}
\Statex
\State Randomly initialize $\theta$, $\phi$, $\psi$
\State Pretrain $G_\theta$ with samples from $p_{data}$
\State Assign the weight from $G_\theta$ to $M_\psi$
    \While {not done}
   \State Sample $\textbf{x}_{r} \sim p_{data}$
   \State Generate $\textbf{x}_{f}$ with $G_\theta$
   \State Compute adversarial loss $\mathcal{L}_{adv}(\theta)$
   \State $\theta' = \theta - \alpha \nabla_\theta \mathcal{L}_{adv}(\theta)$
   \State Compute $M_\psi(\textbf{x}_{r})$ with language model
   \State $g_m = \nabla_\theta \lambda\mathcal{L}_{cot}(M_\psi({\textbf{x}_{r}}), G_{\theta'}(\textbf{x}_{r}))$
    \Comment{Compute meta gradient}
   \State $\theta = \theta - \alpha (\nabla_\theta \mathcal{L}_{adv}(\theta) + \lambda g_m)$
    \Comment {Generator update}
    \State $\phi = \phi - \beta\nabla_\phi \mathcal{L}_{adv}(\phi) $
    \Comment{Discriminator update}
    \State $\psi = \psi - \gamma\nabla_\psi \mathcal{L}_{cot}(\psi)$
    \Comment{Language model update}
    \EndWhile
    \State \Return {Generator $G_\theta$}
\end{algorithmic}
\end{algorithm}


\section{Experiments}
We denote our proposed \underline{meta} \underline{co}operative \underline{t}raining \underline{g}enerative \underline{a}dversarial \underline{n}etworks as {Meta-CoTGAN}. In the experiment section, first, we compare our proposed algorithm with its closest cooperative training counterpart, CoT~\cite{lu2018cot} on the synthetic dataset. Then we show the comparison result between our method and several RL-based and RL-free approaches on  two commonly used real-world text generation datasets: COCO Image Captions~\cite{chen2015microsoft} and EMNLP 2017 WMT News~\footnote{http://statmt.org/wmt17/translation-task.html}.

\paragraph{Implementation Details} We implement our proposed algorithm on top of {RelGAN}~\cite{nie2018relgan}, an RL-free adversarial text generation model that is among the state-of-the-art approaches. Specifically, RelGAN adopts a relational memory to model the long-distance dependencies among the input tokens, and a \textit{gumbel-softmax} relaxation to overcome the non-differentiable issue in the generator training. The relational memory adopts 1 memory slot, multi-head attention with 2 heads, and the attention key size is set to be 512. The language model for cooperative training adopts the identical network architecture as the generator, and the weights for the generator's parameters are assigned to the language model after pretraining. The discriminator adopts multiple representations with size to be 64. We adopt Adam~\cite{kingma2014adam} as the optimization algorithm for updating all the model parameters. The source code of our framework is based on PaddlePaddle\footnote{https://www.paddlepaddle.org.cn/} platform.

\paragraph{Evaluation Metrics} For comparison, we evaluate the models in terms of \textit{sample quality} and \textit{sample diversity} simultaneously. Following most of today's text generation works (e.g., ~\cite{yu2017seqgan,lu2018cot}), the \textit{sample quality} is evaluated by the BLEU score metrics when testified on real datasets, and $\mbox{NLL}_{oracle}$ loss when testified on the synthetic dataset. The $\mbox{NLL}_{oracle}$ loss is defined as the negative log likelihood derived from the target LSTM model for the data generated by $G_\theta$. The \textit{sample diversity} is evaluated in terms of $\mbox{NLL}_{\mbox{gen}}$ loss, which is in the following form:
\begin{equation}\notag
\mbox{NLL}_{\mbox{gen}} = -\mathbb{E}_{\textbf{x}_{1:K}} \mbox{log} P_\theta ( \textbf{x}_1, ..., \textbf{x}_K),
\end{equation}
where the density of the real data is evaluated on the generator model. Thus, models with better sample diversity would have a broader coverage over the real data space and result in lower $\mbox{NLL}_{gen}$ loss. Models that suffer from severe \textit{mode collapse} would no longer represent the real data well and result in higher $\mbox{NLL}_{gen}$ loss.

\paragraph{Baseline Models} To evaluate the efficiency of our proposed approach, we consider {MLE} as well as the RL-based baselines, including {SeqGAN}~\cite{yu2017seqgan}, {RankGAN}~\cite{lin2017adversarial} and {LeakGAN}~\cite{guo2018long}. Also, we compare with the most related RL-free baseline {RelGAN}~\cite{nie2018relgan}. During evaluation, we follow the temperature settings proposed in {RelGAN} and present the results for our method when evaluated with temperature values of $100$ and $1000$, respectively.

\subsection{Synthetic Dataset}
Our first evaluation domain is the synthetic oracle dataset, which is first proposed in~\cite{yu2017seqgan}. The experiment engages a randomly initialized LSTM model as the target model to simulate real-world sequences and generate data from real data distribution. The synthetic experiments are conducted with the sequence length set to be 20. The objective for experimenting in this domain is to compare our proposed method with its closest cooperative training counterpart {CoT}. While these two models adopt same way to train the language model, we investigate on the efficiency of adopting the respective cooperative training losses on the generator model as proposed in these two methods.

\begin{figure}[h!]
\begin{center}
\includegraphics[width=.75\columnwidth]{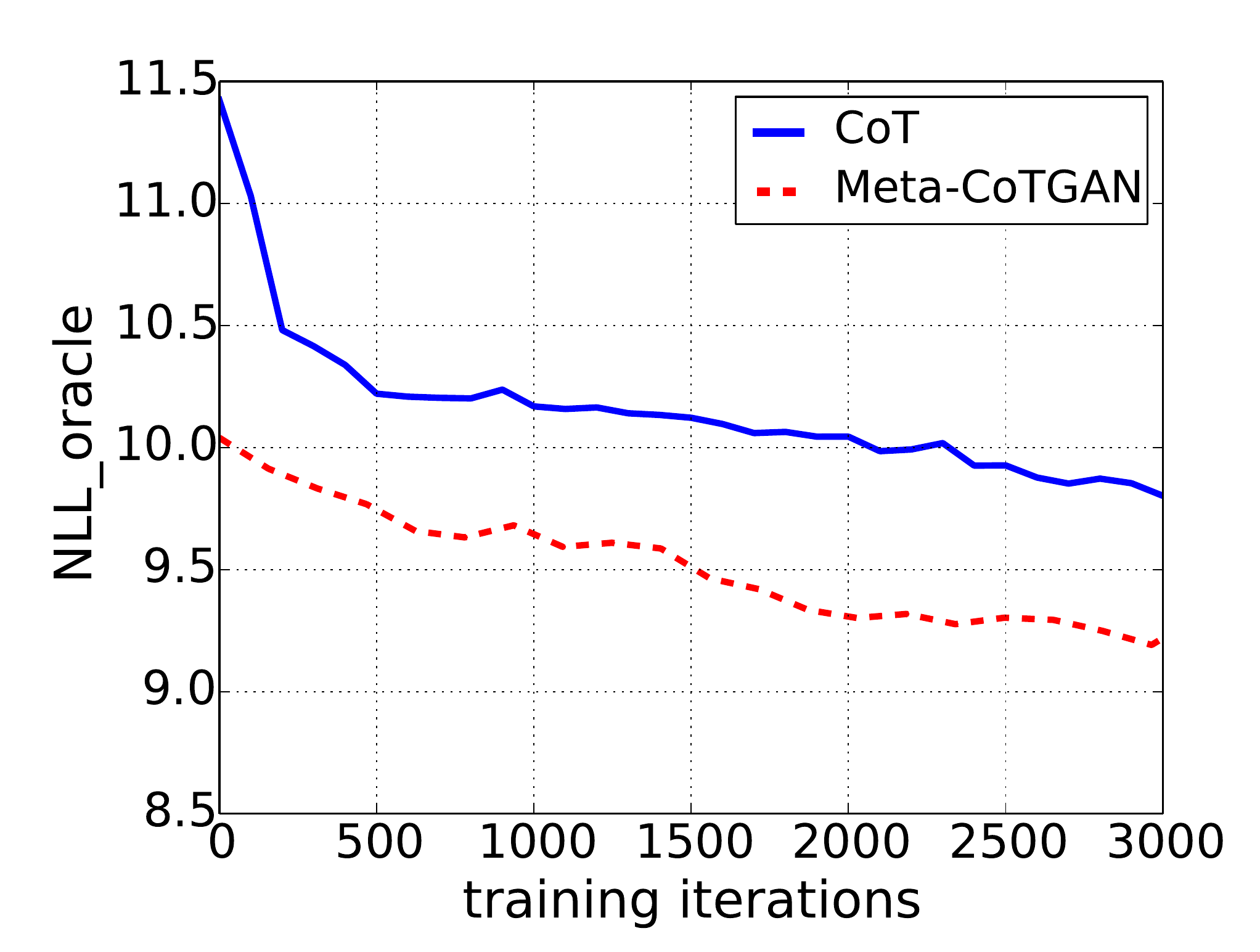}
\end{center}
\vspace{-0.2in}
\caption{Evaluation result on synthetic oracle with length 20 in terms of $\mbox{NLL}_{\mbox{oracle}}$ loss. Overall, our model could converge to significantly better standard than CoT.}
\label{fig:synth}
\end{figure}

We demonstrate the learning curves for $\mbox{NLL}_{\mbox{oracle}}$ loss in Figure~\ref{fig:synth}. Note that CoT takes no pretraining stage and its $\mbox{NLL}_{\mbox{oracle}}$ loss progressively decreases. Our method takes a pretraining stage and the loss decreases in both the pretraining stage and the adversarial training stage. We could notice that upon convergence, the $\mbox{NLL}_{\mbox{oracle}}$ loss for our method is significantly lower than CoT. This demonstrates that the cooperative training mechanism proposed by CoT is not comparable to our method in terms of \textit{sample quality}. We also present the evaluation scores for $\mbox{NLL}_{\mbox{oracle}}$ and $\mbox{NLL}_{\mbox{gen}}$ in Table~\ref{table:synthetic}. When comparing $\mbox{NLL}_{\mbox{gen}}$, our method could achieve much lower loss scale than CoT. This demonstrates that our proposed algorithm convey greater efficiency in preserving the \textit{sample diversity}. Overall, considering the inferior performance and long training time of this model, we do not consider it further in the following real-world dataset experiments.

\begin{table}[h!]
  \centering
  \scalebox{1.0}{%
  \begin{tabular}{l|ll}
    \toprule
   Method   & $\mbox{NLL}_{\mbox{oracle}}$ & $\mbox{NLL}_{\mbox{gen}}$   \\
    \toprule
    CoT & 8.19 & 7.54  \\
    Meta-CoTGAN & \textbf{7.69} & \textbf{6.86} \\
    \toprule
  \end{tabular}
  }
  \caption{Evaluation result on synthetic oracle with sequence length 20. For CoT, we present their \textit{best} score for $\mbox{NLL}_{\mbox{gen}}$.  \label{table:synthetic}
  }
\end{table}

\begin{table*}[t!]
  \centering
  \scalebox{1.0}{%
  \begin{tabular}{l|llll|l}
    \toprule
   Method   & BLEU-2 & BLEU-3 & BLEU-4 & BLEU-5 & $\mbox{NLL}_{\mbox{gen}}$ \\
    \toprule
    MLE & 0.731 & 0.497 & 0.305 & 0.189 & 0.718 \\
    SeqGAN & 0.745 & 0.498 & 0.294 & 0.180 & 1.082 \\
    RankGAN & 0.743 & 0.467 & 0.264 & 0.156 & 1.344 \\
    LeakGAN & 0.746 & 0.528 & 0.355 & 0.230 & 0.679 \\
   RelGAN (100) & 0.849 $\pm$ 0.030 & 0.687 $\pm$ 0.047 & 0.502 $\pm$ 0.048 & 0.331 $\pm$ 0.044 & 0.756 $\pm$ 0.054 \\
   RelGAN (1000) & 0.814 $\pm$ 0.012 & 0.634 $\pm$ 0.020 & 0.455 $\pm$ 0.023 & 0.303 $\pm$ 0.020 & 0.655 $\pm$ 0.048 \\
    \midrule
   Meta-CoTGAN (100) & \textbf{0.858 $\pm$ 0.003} & \textbf{0.692 $\pm$ 0.005} & \textbf{0.518 $\pm$ 0.007} & \textbf{0.363 $\pm$ 0.009} & \textbf{0.578 $\pm$ 0.036} \\
   Meta-CoTGAN (1000) & 0.842 $\pm$ 0.011 & 0.675 $\pm$ 0.019 & 0.502 $\pm$ 0.026 & 0.349 $\pm$ 0.024 & 0.583 $\pm$ 0.028 \\
    \toprule
  \end{tabular}
  }\vspace{-0.1in}
  \caption{Evaluations on COCO Image Captions dataset. For RelGAN and Meta-CoTGAN, the temperature (in parentheses) is set to be 100 and 1000, and results are averaged over 6 runs (random seeds). For $\mbox{NLL}_{\mbox{gen}}$ (last column), the smaller the better.  }
  \label{table:coco}
\end{table*}
\begin{table*}[t!]
  \centering
  \scalebox{1.0}{%
  \begin{tabular}{l|llll|l}
    \toprule
   Method   & BLEU-2 & BLEU-3 & BLEU-4 & BLEU-5 & $\mbox{NLL}_{\mbox{gen}}$ \\
    \toprule
    MLE & 0.768 & 0.473 & 0.240 & 0.126 & 2.382 \\
    SeqGAN & 0.777 & 0.491 & 0.261 & 0.138 & 2.773 \\
    RankGAN & 0.727 & 0.435 & 0.209 & 0.101 & 3.345 \\
    LeakGAN & 0.826 & 0.645 & 0.437 & 0.272 & 2.356 \\
   RelGAN (100) & 0.881$\pm$ 0.013 & 0.705 $\pm$ 0.019 & 0.501 $\pm$ 0.023 & 0.319 $\pm$ 0.018 & 2.482 $\pm$ 0.031 \\
   RelGAN (1000) & 0.837 $\pm$ 0.012 & 0.654 $\pm$ 0.010 & 0.435 $\pm$ 0.011 & 0.265 $\pm$ 0.011 & 2.285 $\pm$ 0.025 \\
   \midrule
   Meta-CoTGAN (100) & \textbf{0.882 $\pm$ 0.014} & \textbf{0.734 $\pm$ 0.017} & \textbf{0.542 $\pm$ 0.016} & \textbf{0.358 $\pm$ 0.015} & 2.299 $\pm$ 0.011 \\
   Meta-CoTGAN (1000) & 0.868 $\pm$ 0.015  & 0.703 $\pm$ 0.014 & 0.500 $\pm$ 0.016 & 0.318 $\pm$ 0.016 &
   \textbf{2.205 $\pm$ 0.053}\\
    \toprule
  \end{tabular}
  }\vspace{-0.1in}
  \caption{Evaluations on EMNLP2017 WMT News dataset. See the caption of Table~\ref{table:coco} for more details.}
  \label{table:emnlp}
\end{table*}

\vspace{-0.2in}
\subsection{COCO Image Captions Dataset}
Our second evaluation domain is the COCO Image Captions dataset. We follow the pre-processing method proposed in~\cite{zhu2018texygen}. The training and testing set consist of 10, 000 sentences respectively. The sentences in COCO have minimum length of 7 and maximum length of 37. The vocabulary size is 4,682.

We present the scores of BLEU-2 to BLEU-5 for measuring \textit{sample quality}, and the $\mbox{NLL}_{\mbox{gen}}$ score for measuring \textit{sample diversity} in Table~\ref{table:coco}. Overall, our method demonstrates significant advantage over all the \textit{sample quality/diversity} metrics. Notably, our method leads to $\mbox{NLL}_{\mbox{gen}}$ loss significantly lower than the other baseline approaches. This indicates that our method could provide an efficient control over the \textit{mode collapse} for the adversarial training and eventually leads to superior \textit{sample diversity}. While decelerating the \textit{mode collapse}, the cooperative training could result in model with better \textit{sample quality} as well.

\begin{table*}[t!]
  \centering
  \scalebox{1.0}{%
  \begin{tabular}{l|llll|l}
    \toprule
   Method   & BLEU-2 & BLEU-3 & BLEU-4 & BLEU-5 & $\mbox{NLL}_{\mbox{gen}}$ \\
    \toprule

   RelGAN (100) & 0.849 $\pm$ 0.030 & 0.687 $\pm$ 0.047 & 0.502 $\pm$ 0.048 & 0.331 $\pm$ 0.044 & 0.756 $\pm$ 0.054  \\
   Meta-CoTGAN (100)  & \textbf{0.858 $\pm$ 0.003} & \textbf{0.692 $\pm$ 0.005 } & \textbf{0.518 $\pm$ 0.007 } & \textbf{0.363 $\pm$ 0.009} & \textbf{0.578 $\pm$ 0.036} \\
   {Meta-CoTGAN}$^{cot-off}$ (100) & 0.824 $\pm$ 0.011 & 0.647 $\pm$ 0.022 & 0.466 $\pm$ 0.028 & 0.315 $\pm$ 0.022 & 0.580 $\pm$ 0.031 \\
   {Meta-CoTGAN}$^{meta-off}$ (100) & 0.835 $\pm$ 0.013 & 0.661 $\pm$ 0.016 & 0.487 $\pm$ 0.016 & 0.338 $\pm$ 0.014 & 0.587 $\pm$ 0.019 \\
   \midrule
   RelGAN (1000) & 0.814 $\pm$ 0.023 & 0.634 $\pm$ 0.020 & 0.455 $\pm$ 0.023 & 0.303 $\pm$ 0.020 & 0.655 $\pm$ 0.048 \\
   Meta-CoTGAN (1000) & 0.842 $\pm$ 0.011 & 0.675 $\pm$ 0.019 & 0.502 $\pm$ 0.026 & 0.349 $\pm$ 0.024 & 0.583 $\pm$ 0.028 \\
   {Meta-CoTGAN}$^{cot-off}$ (1000) & 0.824 $\pm$ 0.007 & 0.643 $\pm$ 0.009 & 0.497 $\pm$ 0.013 & 0.324 $\pm$ 0.015 & 0.582 $\pm$ 0.017 \\
   {Meta-CoTGAN}$^{meta-off}$ (1000) &  0.817 $\pm$ 0.021 & 0.638 $\pm$ 0.027 & 0.465 $\pm$ 0.025 & 0.319 $\pm$ 0.018 & 0.589 $\pm$ 0.022 \\

    \toprule
  \end{tabular}
  }\vspace{-0.1in}
\caption{Ablation study result on COCO Image Captions dataset. We evaluate our proposed model when the cooperative training part and meta optimization have been turned off, respectively. Reported scores are derived from 6 random seeds. }
\label{table:abl-cot}\vspace{-0.1in}
\end{table*}

\begin{figure}[h!]
\begin{center}
\vspace{-0.25in}
\includegraphics[width=.75\columnwidth]{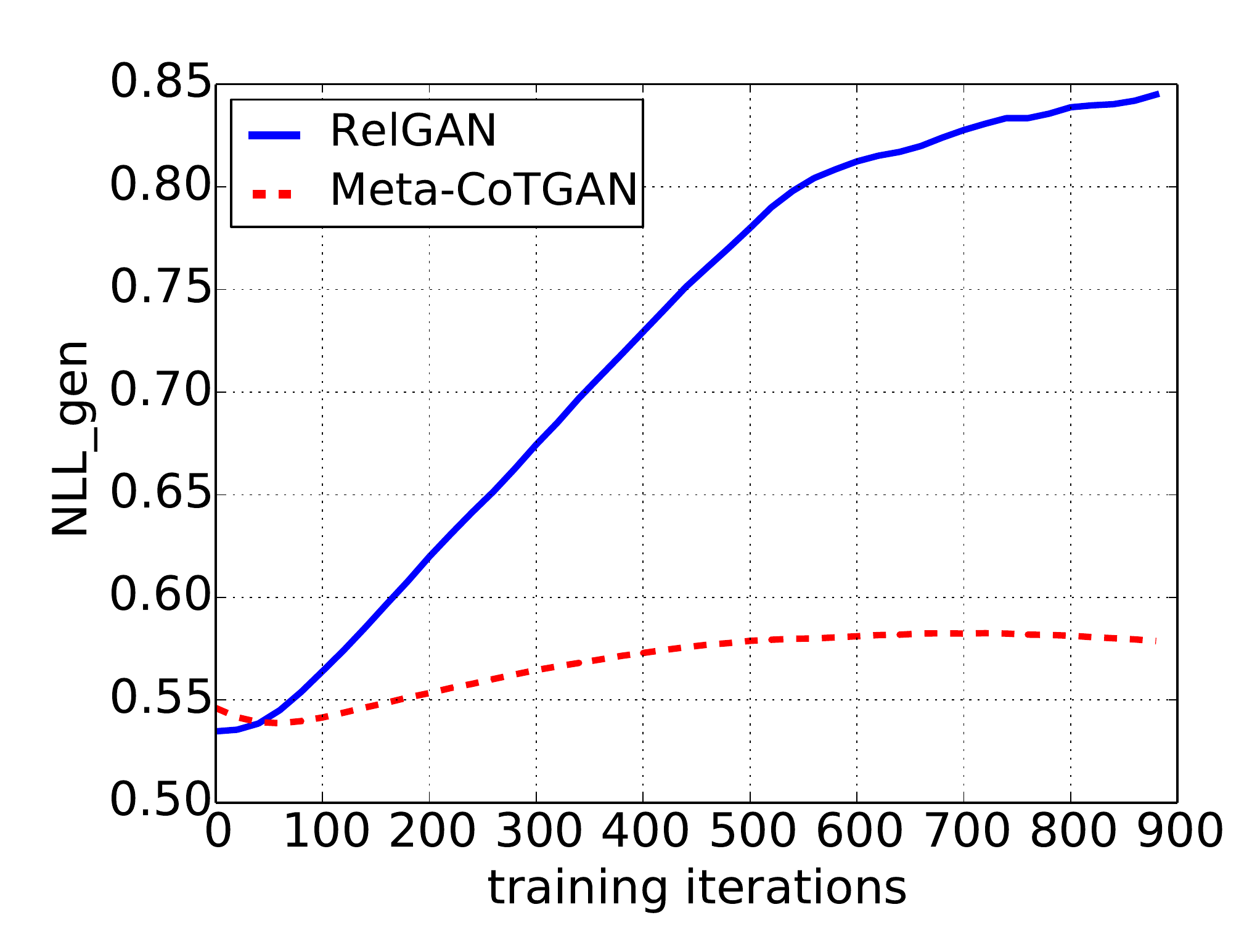}
\vspace{-0.1in}
\includegraphics[width=.75\columnwidth]{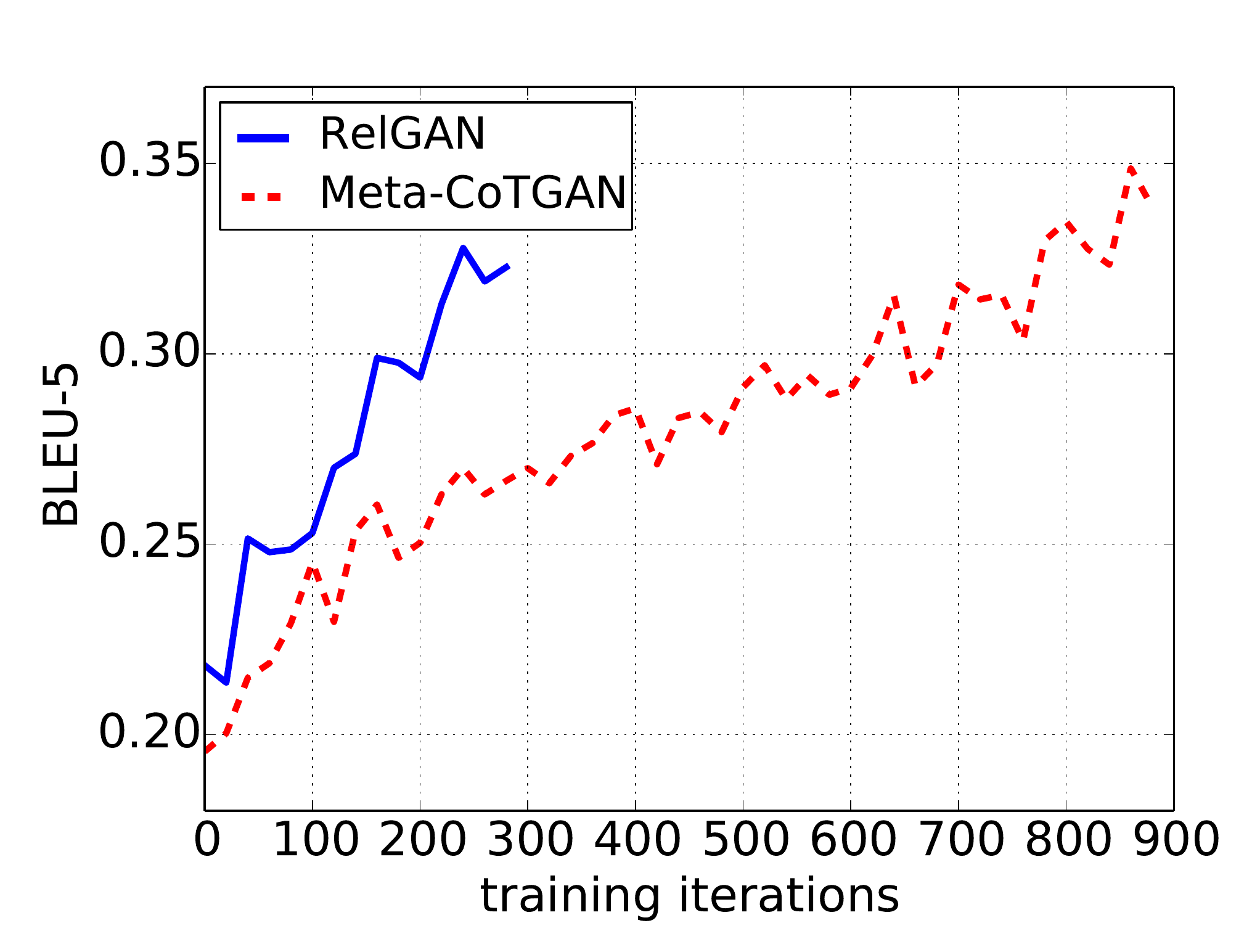}
\end{center}
\vspace{-0.2in}
\caption{We demonstrate the quality-diversity trade-off for our method as well as the baseline {RelGAN} on COCO Image Captions dataset. Our model progressively achieves better BLEU-5 score than {RelGAN} with an apparently slow progress for \textit{mode collapse}. The BLEU-5 for {RelGAN} is plotted up to the point when its corresponding $\mbox{NLL}_{\mbox{gen}}$ loss reaches its reported standard. Otherwise, the BLEU-5 score becomes no more meaningful since the model has turn into severe \textit{mode collapse} (i.e., generating repeated sentences). }
\label{fig:coco}\vspace{-0.3in}
\end{figure}

To further validate this, we present the learning curves for the \textit{sample diversity} metric and BLEU-5 as a representative \textit{sample quality} metric in Figure~\ref{fig:coco}. We could observe that the $\mbox{NLL}_{\mbox{gen}}$ for {RelGAN} would fast go up, which is a sign of \textit{mode collapse}. However, that for {MetaCoTGAN} progresses rather slowly. It shows that our proposed method could efficiently decelerate \textit{mode collapse} and control the $\mbox{NLL}_{\mbox{gen}}$ loss from explode. When investigating on the \textit{sample quality} metric, we could observe the BLEU-5 score for {RelGAN} would go up faster than {MetaCoTGAN}. But eventually, our model could achieve a significantly higher standard than {RelGAN}.
Also, we observe that when $\mbox{NLL}_{\mbox{gen}}$ for {RelGAN} explode (e.g., after 400 epochs), the repeat rate is rather high and therefore the generator just becomes useless. However, our method could preserve much better diversity.
Also, we observe from the generated real sentences that our model could generate quite long sentences, while most of the GAN models that fall short~\cite{caccia2018language}.

\subsection{EMNLP2017 WMT News Dataset}
Our third evaluation domain is the EMNLP2017 WMT News dataset. The size of this dataset is much larger than Image COCO, involving a training set of 270,000 sentences. The testing set consists of 10,000 sentences. The sentences have maximum length of 51. The vocabulary size is 5,255.

The results for EMNLP dataset are presented in Table~\ref{table:emnlp}. We can see that our proposed method consistently outperforms all  baselines in terms of all the BLEU metrics and $\mbox{NLL}_{\mbox{gen}}$. Under the temperature setting of 100, our method outperforms the strong {RelGAN} baseline by $0.041/0.039$ on BLEU-4/BLEU-5. Noticeably, the best BLEU scores for our method are obtained when the $\mbox{NLL}_{\mbox{gen}}$ loss is at a significantly lower level than {RelGAN}. This indicates that by conducting cooperative training, we could derive generator model with better \textit{sample quality} and \textit{sample diversity} simultaneously. Moreover, it shows that our method could robustly perform well in rather challenging and diverse real-world datasets like EMNLP. Meanwhile, the performance of our method is quite robust, consistently outperforming {RelGAN} under both temperature settings, over all the evaluation metrics.
By investigating through the generated real samples, we observe that the generated sentences convey rather diverse semantics and the output consists of considerably long sentences, unlike the conventional adversarial text generators that would shortly fall to the phase of generating short and repeated sentences.

\subsection{Ablation Study}
\subsubsection{Impact of Cooperative Training Language Model}
We demonstrate the impact of using an online updated language model to conduct our proposed cooperative training process. To this end, a direct comparison is to use a pretrained language model not updated with cooperative training. We denote such baseline as {Meta-CoTGAN}$^{cot-off}$. We demonstrate the result on COCO Image Captions dataset in Table~\ref{table:abl-cot}. We could observe that when online update to the language model is turned off, the model still preserve comparable \textit{sample diversity} in terms of $\mbox{NLL}_{\mbox{gen}}$, since the cooperative training loss is still employed on the real data. However, under both temperature setting, the \textit{sample quality} metrics could not perform as well as the full set of the proposed method. This shows that it is beneficial to update the language model jointly with the generator to let it offer a smoothly chanting target distribution to the generator.

\subsubsection{Impact of Meta Optimization}
We also evaluate the impact of the meta optimization setup. To this end, we compare our approach with a principled way of engaging the cooperative training loss for optimizing the generator parameters, which is proposed in the form of linearly summing up the adversarial loss and the cooperative training loss in a weighted manner, i.e., $\mathcal{L}_{adv}(\theta) + \lambda \mathcal{L}_{\cot}(\theta)$. We denote such baseline as {Meta-CoTGAN}$^{meta-off}$. The results are shown in Table~\ref{table:abl-cot}. Overall, {Meta-CoTGAN}$^{meta-off}$ obtain comparable scores for $\mbox{NLL}_{\mbox{gen}}$. However, its performance in terms of the \textit{sample quality} metrics is still much inferior than using full set of solution. Thus, we could conclude that meta optimization is an important ingredient for balancing the quality-diversity trade-off. Intuitively, our proposed meta optimization set-up offers an efficient way to ensure the generator parameters  \textit{after} the adversarial update would decelerate from \textit{mode collapse}, which is critical to derive the superior performance.

\section{Conclusion and Discussion}
We propose a meta cooperative training approach to facilitate the training of adversarial text generation models. Our method utilizes a cooperatively trained language model to effectively decelerate the \textit{mode collapse} of adversarial training via distilling the prediction output distribution of the language model over the real data to the adversarial generator model. We evaluate our proposed method in both synthetic dataset and two real-world datasets with sequence length at a range from 7 to 51. As a result, our proposed method could consistently outperform the baseline algorithms on \textit{sample quality} metrics and \textit{sample diversity} metric simultaneously. Our proposed approach is general and could be promising to work with distinct RL-based or RL-free adversarial text generation algorithms as long as they face the issue of \textit{mode collapse}. Our future work would be to apply meta cooperative training on more emerging RL-based/free GAN models.

\bibliographystyle{aaai20}
\bibliography{aaai}

\end{document}